\documentclass[runningheads,a4paper]{llncs}
\usepackage[top=2.5cm, bottom=2.5cm, left=2.5cm, right=2.5cm]{geometry}
\usepackage{amsmath}
\usepackage{amssymb}
\usepackage{graphicx}
\usepackage{booktabs}
\usepackage{listings}
\usepackage{booktabs} 
\usepackage{graphicx} 
\usepackage{caption}  
\usepackage{multirow}

\usepackage{hyperref}
\hypersetup{unicode=true}
\hypersetup{pdfborder={0 0 1}}

\hypersetup{
    linkbordercolor={1 1 1}, % White border color (effectively invisible)
}

\usepackage{xcolor}

\AtBeginDocument{%
    \hypersetup{
        citebordercolor={1 0 0} % Blue border for citations, adjust to your preference
    }
}
\let\oldhyperref\hyperref
\renewcommand{\hyperref}[2][]{\oldhyperref[#1]{\textbf{#2}}}
\usepackage[noadjust,nocompress]{cite}
\usepackage{caption}
\usepackage{subcaption}

\usepackage{url}
\usepackage{color}
\usepackage{epsfig}
\usepackage{float}
\usepackage{bm}

\makeatletter
\renewcommand{\paragraph}{%
  \@startsection{paragraph}{4}%
  {\z@}{3.25ex \@plus 1ex \@minus .2ex}{-1em}%
  {\normalfont\normalsize\bfseries}%
}
\makeatother

\usepackage[nomargin,inline,marginclue,draft]{fixme}
\usepackage{cleveref}
\fxsetup{status=draft}
\fxsetup{theme=color, mode=multiuser} \FXRegisterAuthor{me}{ame}{\color{red}Me}

\title{Developing Foundation Models for Universal Segmentation from 3D Whole-Body Positron Emission Tomography}

\titlerunning{Developing Foundation Models for Universal 3D PET Segmentation}

\author{ 
    {\small Yichi Zhang\inst{1,2,\dag}}
    \and
    {\small Le Xue\inst{2,3,\dag}}
    \and
    {\small Wenbo Zhang\inst{2,4}}
    \and
    {\small Lanlan Li\inst{4}}
    \and
    {\small Feiyang Xiao\inst{1,2}}
    \and
    {\small Yuchen Liu\inst{1,2}}
    \and
    \\
    {\small Xiaohui Zhang \inst{5}}    
    \and
    {\small Hongwei Zhang\inst{1,2}}
    \and
    {\small Shuqi Wang\inst{4}}   
    \and
    {\small Gang Feng \inst{6}}    
    \and
    {\small Liling Peng \inst{6}}    
    \and
    {\small Xin Gao \inst{6}}
      \and
    {\small Yuanfan Xu \inst{7}} 
    \and
     \\
    {\small Yuan Qi \inst{1,2,8}}
    \and
    {\small Kuangyu Shi \inst{9}}
    \and
    {\small Hong Zhang \inst{5,*}}    
    \and
    {\small Yuan Cheng \inst{1,2,*}}
    \and
    {\small Mei Tian \inst{2,3,4,*}}    
    \and
    {\small Zixin Hu \inst{1,2,*}}
    }

 \authorrunning{Zhang et al.}

    \institute{{\small{$^{1}$ Artificial Intelligence Innovation and Incubation Institute, Fudan University, Shanghai, China. \\ \quad $^{2}$ Shanghai Academy of Artificial Intelligence for Science, Shanghai, China. \\ \quad $^{3}$ Department of Nuclear Medicine and PET Center, Huashan Hospital, Fudan University, Shanghai, China. \\ \quad $^{4}$ Human Phenome Institute, Fudan University, Shanghai, China. \\  \quad $^{5}$ Department of Nuclear Medicine, Zhejiang Cancer Hospital, Hangzhou Institute of Medicine, Chinese Academy of Sciences, Hangzhou, China. \\ \quad $^{6}$ Shanghai Universal Medical Imaging Diagnostic Center, Shanghai, China. \\ \quad $^{7}$ Hangzhou Universal Medical Imaging Diagnostic Center, Hangzhou, China.  \\  \quad $^{8}$ Department of Information and Intelligence Development, Zhongshan Hospital, Fudan University, Shanghai, China. \\ \quad $^{9}$ Department of Nuclear Medicine, Inselspital, University of Bern, Switzerland. }
   }
    ~
    \\
    ~
        \\
\small{\dag} These authors contributed equally to this work.
\small{*} Corresponding authors.
}

\begin{document}
    \mainmatter
    \maketitle

\setcounter{footnote}{0} 
\begin{abstract}

Positron emission tomography (PET) is a key nuclear medicine imaging modality that visualizes radiotracer distributions to quantify in vivo physiological and metabolic processes, playing an irreplaceable role in disease management.
Despite its clinical importance, the development of deep learning models for quantitative PET image analysis remains severely limited, driven by both the inherent segmentation challenge from PET’s paucity of anatomical contrast and the high costs of data acquisition and annotation. 
To bridge this gap, we develop generalist foundational models for universal segmentation from 3D whole-body PET imaging.
We first build the largest and most comprehensive PET dataset to date, comprising 11,041 3D whole-body PET scans with 59,831 segmentation masks for model development. Based on this dataset, we present SegAnyPET, an innovative foundational model with general-purpose applicability to diverse segmentation tasks. Built on a 3D architecture with a prompt engineering strategy for mask generation, SegAnyPET enables universal and scalable organ and lesion segmentation, supports efficient human correction with minimal effort, and enables a clinical human-in-the-loop workflow. Extensive evaluations on multi-center, multi-tracer, multi-disease datasets demonstrate that SegAnyPET achieves strong zero-shot performance across a wide range of segmentation tasks, highlighting its potential to advance the clinical applications of molecular imaging.
\end{abstract}

    %\begin{refsegment}  % Begin a new reference section for main content

    %\keywords{diagnosis, machine learning }
    % =========================
    % main part of the document
    % =========================
    \section*{Introduction}

Positron emission tomography (PET) is a powerful molecular imaging modality that offers a unique perspective on a patient's health by visualizing radiotracer distribution to reveal physiological processes \cite{phelps2000positron}.
Unlike other imaging modalities that capture anatomical structures, PET can detect early disease onset by revealing metabolic changes before physical alterations are visible on structural imaging like computed tomography (CT) and magnetic resonance imaging (MRI) \cite{basu2011fundamentals}. This functional imaging capability makes it an invaluable tool for modern medical diagnostics, particularly in oncology and neurology \cite{zhu2016multi,peng202318f,xue202418f,schwenck2023advances,sun2022identifying}.
Accurate volumetric segmentation of organs and lesions from PET volumes is essential for comprehensive and quantitative multi-systemic analysis of interactions between different organs and pathologies, so as to assist in clinical workflows \cite{zaidi2010pet,bagci2013joint,yousefirizi2021toward}.
However, PET imaging inherently lacks high-contrast anatomical boundary information, making the extraction of precise anatomical features for organ and lesion delineation far more technically challenging than from conventional anatomical modalities \cite{wang2025robust}.
Manual delineation of PET volumes remains time-consuming, observer-dependent and poorly reproducible in routine clinical practice, creating an urgent demand for robust high-precision segmentation solutions.

Driven primarily by advances in deep learning, considerable progress has been made on the problem of volumetric segmentation \cite{azad2024medical}. Typically, these deep models are developed by collecting imaging data and annotating the corresponding organs or tumors for training, thereby enabling the segmentation of these trained targets.
The emergence of numerous segmentation datasets \cite{AbdomenCT-1K,ma2022fast,ma2024unleashing,li2024abdomenatlas,wasserthal2023totalsegmentator} has significantly expanded the scope of target organs and substructures.
The growing availability of large-scale, multi-center datasets is increasingly driving the development of generalizable models, enabling robust performance across diverse clinical environments.
However, these breakthroughs in volumetric segmentation have predominantly focused on CT or MRI, while research on PET imaging remains far behind. A significant obstacle contributing to this gap stems from the high costs associated with the acquisition and annotation of PET images. The inherent limitations of PET imaging, such as low spatial resolution and the presence of partial volume effects, further hinder the development and validation of PET segmentation \cite{iantsen2021convolutional}.
Although there are publicly available PET segmentation datasets \cite{AutoPET,hecktor}, they are largely restricted to highly specific oncological tasks, lacking the comprehensive, whole-body coverage of diverse organs and lesions.
This narrow focus presents a critical limitation, as deep models are tailored to specific segmentation targets that align with the training data during model development. 
Consequently, they often lack generalization capability across diverse tasks, necessitating additional training with annotations when adapted to new segmentation targets.
This limitation restricts the utility of existing models when confronting the complex demands of real-world clinical scenarios, which inherently require the simultaneous segmentation and quantitative evaluation of diverse targets \cite{ma2023towards}. 

Recent developments of foundation models holds promise for providing an universal solution \cite{moor2023foundation,willemink2022toward}. Foundation models are large-scale deep models trained on large amounts of data with remarkable generalization abilities across a wide range of downstream tasks \cite{zhou2023foundation,zhang2025echo,gao2025lung,tak2026generalizable,liang2026foundation,blankemeier2026merlin}.
For image segmentation, the introduction of the Segment Anything Model (SAM) \cite{SAM} has attracted significant attention as a universal foundation model capable of producing fine-grained segmentation for arbitrary targets using prompts such as points.
Following its success in natural images, several studies have attempted to adapt SAM to the medical domain \cite{SAM4MIS}. However, these medical foundation models are predominantly constructed upon microscopy images whose imaging principles closely resemble those of natural images \cite{marks2025cellsam,archit2025segment}, or structural radiological modalities such as CT and MRI \cite{MedSAM,SAM-Med3D,du2024segvol,zhao2025foundation}. Constrained by the limited availability of comprehensively annotated cohorts, functional PET imaging is notably absent, leaving its distinct characteristics largely overlooked.
As these foundation models are primarily optimized to delineate the well-defined anatomical boundaries inherent in structural scans, they lack the specific representational capacity required to interpret the diffuse, metabolism-driven signal distributions and low spatial resolution typical of PET. Although promising performance has been reported on other segmentation tasks, the universal segmentation paradigm fails to effectively generalize to PET tasks, leaving a critical void in foundation models capable of reliably parsing functional imaging \cite{zhang2026uncovering}.

To overcome these challenges, we introduce SegAnyPET, a foundational model for universal volumetric segmentation from whole-body PET imaging, which utilizes the design of SAM for universal promptable segmentation with an extension to 3D architecture to fully utilize the inter-slice context information of PET volumes.
To enable model development, we construct PETWB-Seg11K, a large-scale, multi-center, and multi-device dataset of 11,041 whole-body 3D PET images and 59,831 segmentation masks.
We conduct a large-scale study to evaluate SegAnyPET on both internal cohorts and external cohorts on unseen target organs, modality variations, and novel radiotracer uptake patterns.
Experimental results demonstrate that SegAnyPET consistently outperforms the state-of-the-art segmentation foundation model with with the same or less prompting efforts.
Besides, SegAnyPET achieves performance on par with, or even surpassing well-designed task-specific models that were trained on training-visible targets, while maintaining strong generalization ability on unseen targets.
We also demonstrate that the segmentation results from SegAnyPET can be efficiently applied to downstream applications.
This work represents a major step forward in creating generalizable and clinically useful AI models for advancing the applications to molecular imaging.

\begin{figure*}[htbp]
    \centering
    \includegraphics[width=\linewidth]{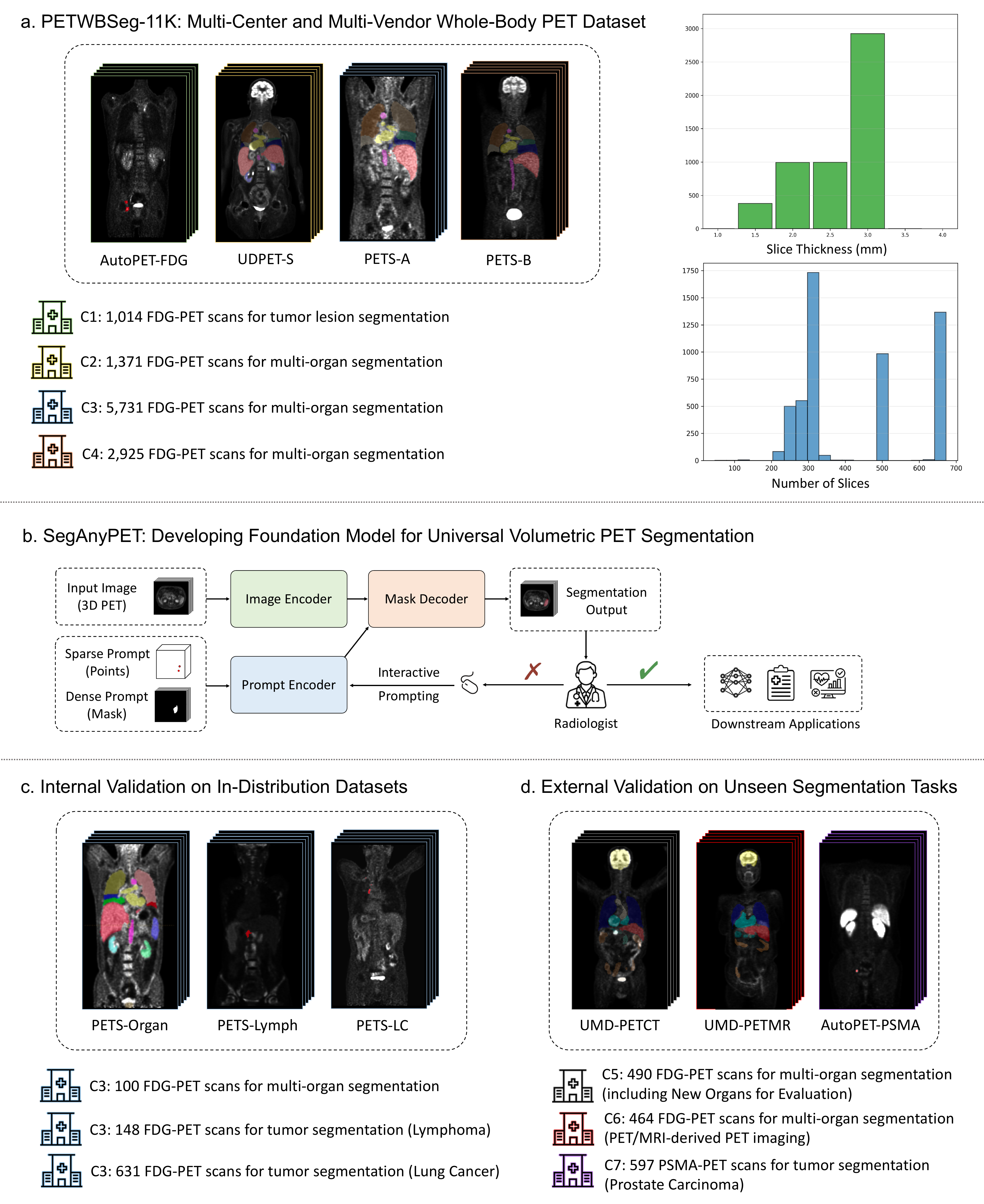}  
    \caption{\textbf{Figure 1 : Overview of dataset construction, model development and validation in our study.} 
    (a) We construct PETWB-Seg11K, a large-scale whole-body PET scans from multi-tracer, multi-vendor, and multi-disease cohorts, with diverse heterogeneity in data distribution.
    (b) We develop SegAnyPET, a foundation model for universal volumetric PET segmentation. It takes volumetric PET input, encodes features via Image Encoder, and combines with sparse or dense prompts encoded by Prompt Encoder, then outputs segmentation through Mask Decoder. SegAnyPET supports a human-in-the-loop workflow with interactive prompting for rectification of segmentation results.
    (c) Comprehensive evaluation on in-distribution segmentation tasks validates the consistent and reliable performance of SegAnyPET on data with similar distribution to the training set.
    (d) Extensive assessment on out-of-distribution segmentation tasks demonstrates the generalization capability of SegAnyPET on unseen data scenarios in cross-center and cross-tracer settings.}
    \label{fig:overview}
\end{figure*}

    \section*{Results}
        \subsection*{Developing foundation models for universal volumetric PET segmentation}

To develop a foundation model for universal segmentation from functional PET images, we introduce PETWB-Seg11K, a large-scale dataset comprising 11,041 whole-body 3D PET volumes with 59,831 segmentation masks, as illustrated in Figure \ref{fig:overview}(a).
PETWB-Seg11K integrates whole-body PET imaging from two open-source datasets \cite{AutoPET,xue2025udpet} and three private cohorts with comprehensive annotation covering a variety of organs and lesions across the entire body.
PETWB-Seg11K captures substantial real-world variability in scanner vendors, acquisition protocols, anatomical coverage, slice numbers, slice thickness, and disease presentations, closely reflecting clinically encountered PET imaging conditions.
Such heterogeneity in data distribution is critical for developing robust foundation models that can generalize beyond the training domain and remain resilient to out-of-distribution scenarios commonly observed in clinical practice.
Supplementary Table 1 provides more detailed information about the dataset.
To support rigorous model evaluation, we construct internal validation sets for in-domain testing, which consist of previously unseen cases acquired from the same clinical centers and imaging protocols as the training data.
In addition, we incorporate two external validation sets from independent centers with unseen cancer types, including PET imaging using different tracers (i.e. PSMA-PET), none of which are observed during model development. This evaluation protocol enables an unbiased and stringent assessment of model generalization under out-of-distribution settings, including shifts in data sources and radiotracer characteristics.
To the best of our knowledge, PETWB-Seg11K is the largest and most comprehensive whole-body PET segmentation dataset to date, substantially exceeding existing PET datasets in both scale and diversity, and providing a solid foundation for advancing universal PET segmentation models.

Motivated by the recent advances in segmentation foundation models, we introduce SegAnyPET, a foundation model to fulfill the role of foundation model for functional PET imaging which enables universal volumetric segmentation.
Leveraging the PETWB-Seg11K dataset for training allows SegAnyPET to capture the intricate functional patterns and generalized representations intrinsic to PET imaging.
Unlike conventional deep models that are trained for a fixed set of annotated categories, SegAnyPET is designed to flexibly segment any organ, tissue, or lesion, including structures not explicitly labeled during training.
As illustrated in Figure \ref{fig:overview}(b), SegAnyPET adopts a holistic 3D architecture that directly leverages volumetric spatial cues from whole-body PET. The framework comprises three key components: an image encoder, a prompt encoder, and a mask decoder. The image encoder extracts discrete 3D feature embeddings from the input PET volume, while the prompt encoder converts user-provided inputs including sparse prompts (e.g., points) or dense prompts (e.g., coarse masks) into compact prompt embeddings through fixed positional encoding and adaptive prompt-specific embedding layers.
These representations are fused and decoded via the mask decoder, which upsamples the combined features and passes them through a multi-layer perceptron to produce the final segmentation output.
This design leverages point prompts to facilitate fast and efficient interaction for 3D imaging. Furthermore, the incorporation of mask prompts enables the model to perform iterative refinement of initial predictions through human guidance.

SegAnyPET fundamentally differs from conventional task-specific deep models that are trained to segment a fixed set of predefined anatomical structures or lesion types.
In contrast, by incorporating prompt-based interaction, SegAnyPET enables universal, user-guided promptable segmentation, allowing radiologists to dynamically specify regions of interest at inference time without retraining or architectural modification.
This design is particularly valuable in whole-body PET interpretation, where lesions may exhibit high inter-patient variability, ambiguous boundaries, or rare manifestations that fall outside predefined label spaces. When radiologists identify missed lesions, inaccurate boundaries, or clinically relevant targets not covered by the original training taxonomy, they can simply provide additional positive or negative point prompts to iteratively refine the segmentation results. This interactive, human-in-the-loop workflow aligns closely with routine clinical practice, in which expert knowledge is continuously applied to guide image interpretation.

\begin{figure*}[htbp]
    \centering
    \includegraphics[width=\linewidth]{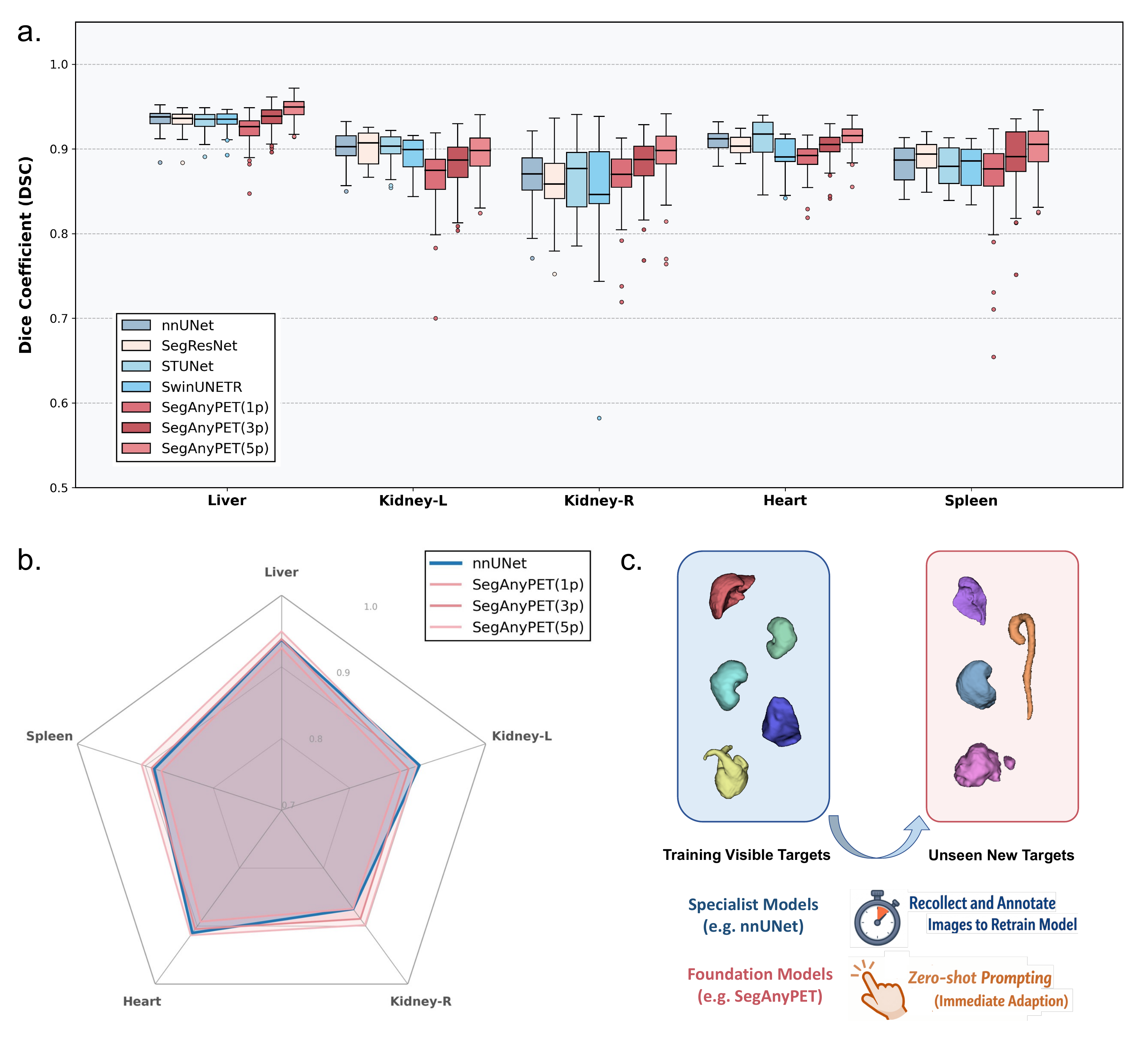}  
    \caption{\textbf{Figure 2: Comparison of SegAnyPET with task-specific segmentation models on internal evaluation for multi-organ segmentation. }
    (a) Quantitative performance comparison with four representative task-specific segmentation models. The center line within each box indicates the median value; the bottom and top bounds indicate the $25_{th}$ percentiles and $75_{th}$ percentiles, respectively. Outlier classes are plotted as individual dots.
    (b) Performance comparison between SegAnyPET and the most competitive task-specific model nnU-Net on five training visible target organs.
    (c) Compared with task-specific models that are trained for the segmentation of training visible targets, SegAnyPET retains the generalization capability to new targets. In contrast, additional annotation and training are required for task-specific models when adapted to training invisible targets.}
    \label{fig:specialist}
\end{figure*}

To accommodate diverse clinical needs in whole-body PET interpretation, we develop two SegAnyPET variants with different levels of specialization.
The SegAnyPET model serves as a general-purpose segmentation foundation model, trained on the full dataset to provide broad coverage across organs and lesions with strong generalization capability.
For clinical scenarios that demand enhanced lesion delineation, particularly for small heterogeneous abnormalities, we further introduce SegAnyPET-Lesion, a specialized variant obtained by fine-tuning SegAnyPET on lesion-centric training data. This specialization improves sensitivity and boundary accuracy for challenging lesions, while retaining the prompt-based flexibility of the model.
Together, these variants enable users to select either a broadly applicable model for routine whole-body analysis or a lesion-focused model tailored to more demanding oncological assessment tasks.

\subsection*{Comparison with state-of-the-art task-specific segmentation models}

We compare SegAnyPET with representative task-specific specialist models, including nnUNet \cite{isensee2020nnunet}, STUNet \cite{huang2023stunet}, SwinUNETR \cite{hatamizadeh2021swin} and SegResNet \cite{segresnet}. These models are designed for fully supervised, end-to-end segmentation on predefined targets and do not support prompt-based interaction. 
To ensure a fair and task-aligned comparison, we train a group of task-specific models using all the organ annotations (C2–C4) within the PETWB-Seg11K dataset. This allows the specialist models to focus exclusively on their intended task of organ segmentation.
Notability, to establish a rigorous baseline, all these models are implemented within the standardized nnUNet framework, adopting its automated data preprocessing, resampling, and augmentation pipelines. 
In contrast, SegAnyPET is trained and evaluated on the entire dataset as a universal promptable segmentation foundation model, handling both lesion and organ segmentation tasks. This approach ensures flexibility and task adaptability without retraining, providing a comprehensive solution across multiple segmentation tasks.

Experimental results in Figure \ref{fig:specialist}(a) show that, despite being proposed a few years ago, nnUNet remains competitive compared with more recent architectural variants. This observation is consistent with prior studies and highlights the robustness of nnUNet when sufficient task-specific supervision is available \cite{isensee2024nnu}.
In contrast, despite being trained as a general-purpose segmentation model rather than a task-specific specialist, SegAnyPET achieves comparable or superior performance to these specialist models on both tumor- and organ-focused evaluation settings.
Notably, SegAnyPET demonstrates competitive segmentation accuracy without the need for per-task retraining (Figure \ref{fig:specialist}(b)), while maintaining the ability to generalize to unseen segmentation targets and lesion types through prompt-based interaction (Figure \ref{fig:specialist}(c)). 
This flexibility contrasts with task-specific models, which are inherently constrained to the predefined label sets encountered during training, limiting their adaptability to new or unseen targets. This allows a single SegAnyPET model to effectively replace multiple task-specific networks, achieving specialist-level accuracy while providing substantially greater flexibility and clinical applicability.
Overall, these results demonstrate that SegAnyPET performs on par with state-of-the-art task-specific models in in-domain evaluation, while also offering unique advantages in robust generalization and interactive refinement for unseen tasks, which is crucial for practical applications in real-world whole-body PET analysis.

\subsection*{Quantitative and qualitative analysis of universal promptable PET segmentation}

\begin{figure*}[tbp]
    \centering
    \includegraphics[width=\linewidth]{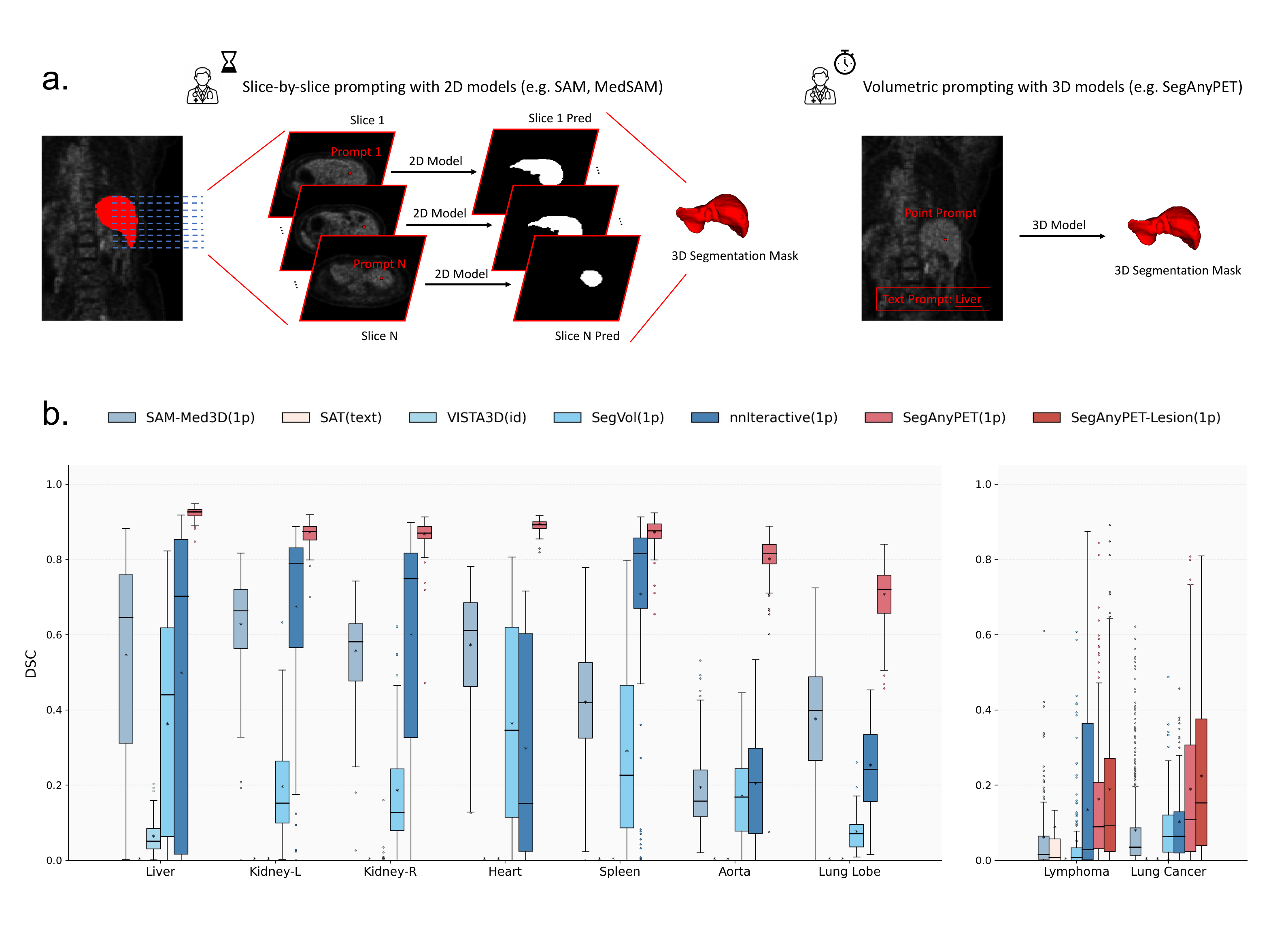}  
    \caption{\textbf{Figure 3: Quantitative and qualitative evaluation of SegAnyPET with other foundation models for promptable PET segmentation.} 
    (a) Illustration of slice-wise and volumetric prompting strategies using 2D and 3D segmentation foundation models. For 2D models (e.g., SAM, MedSAM), the PET volume is decomposed into individual slices, where point prompts are provided on selected slices and the resulting 2D predictions are aggregated to form a 3D segmentation. In contrast, 3D foundation models (e.g., SegAnyPET) directly operate on the entire volumetric input and perform end-to-end volumetric segmentation.
    (b) Quantitative performance comparison of different promptable segmentation foundation models across multiple internal PET segmentation tasks. The DSC score is calculated in a volume-wise manner.}
    \label{fig:prompting}
\end{figure*}

We evaluate SegAnyPET through both in-distribution and out-of-distribution validation for universal promptable segmentation, and conduct comprehensive comparisons with representative 3D general-purpose segmentation foundation models. Specifically, we benchmark SegAnyPET against state-of-the-art promptable segmentation models including SAM-Med3D \cite{SAM-Med3D}, SegVol \cite{du2024segvol}, SAT \cite{zhao2025large}, nnIteravtive \cite{isensee2025nninteractive} and VISTA3D \cite{he2025vista3d}, all of which are claimed to be capable of general-purpose medical segmentation.
As shown in Figure \ref{fig:prompting}(a), all evaluations are performed at the 3D volumetric level to ensure a consistent and clinically meaningful comparison. For 2D models, axial slices containing the target organ or lesion are first identified, and prompt-based segmentation is conducted on each slice independently. The resulting 2D predictions are then aggregated to form the final 3D segmentation. In contrast, 3D foundation models directly take the entire PET volume as input and generate volumetric segmentation outputs in an end-to-end manner.
For models driven by textual prompts, we provide the semantic category of the target structure to be segmented following their original design protocols. This unified evaluation protocol enables a fair assessment of different prompt modalities and model dimensionalities.

Figure \ref{fig:prompting}(b) presents quantitative comparison of 3D models across multiple organ and lesion segmentation tasks for internal evaluation.
While numerous attempts have been made to extend the success of promptable foundation model to medical image segmentation, these adaptations face significant limitations when generalizing to PET imaging.
A primary reason is that most purportedly general-purpose medical foundation models are predominantly trained and evaluated on structural imaging modalities such as CT, which benefit from large-scale dataset availability (See Figure \ref{fig:dataused} for details). Consequently, PET imaging has been largely overlooked in model development due to data scarcity.
A highly intuitive observation from the results is the complete failure of text prompt-based models, such as SAT, which yield DSC approaching zero for organ segmentation tasks. This profound underperformance indicates that the cross-modal alignment between textual descriptions and visual features in these models is heavily overfitted to the anatomical structures seen in structural imaging. Consequently, they lack the generalization capacity required to interpret the complex, functional uptake patterns and whole-body metabolic distributions characteristic of PET imaging.
Furthermore, PET exhibits a substantial domain gap compared to structural imaging modalities, characterized by low signal-to-noise ratio and ill-defined boundaries, which collectively render segmentation significantly more challenging.
While models dependent on positional point prompts like SAM-Med3D and nnIteractive deliver slightly better segmentation outcomes, their overall performance remains inadequate. By explicitly providing spatial coordinates, these point prompts help anchor the model's attention to target organs or lesions. Nevertheless, the overall efficacy remains highly constrained. While the positional priors guide localization, the foundational features extracted by these models pre-trained on structural data fail to accurately delineate PET imaging.
In contrast, SegAnyPET consistently outperforms state-of-the-art promptable segmentation foundation models across all evaluated tasks. We attribute this improvement primarily to the fact that SegAnyPET is trained on large-scale and diverse PET data, enabling it to efficiently learn PET-specific representations. As existing foundation models are predominantly trained on structural imaging modalities, their appearance statistics and anatomical priors differ substantially from functional PET images, leading to degraded performance when directly transferred to PET.

\subsection*{Generalization evaluation on unseen data distribution and different radiotracers}

We further evaluate the generalization capability of SegAnyPET using three external validation datasets that are entirely different from model development. These experiments are designed to assess model robustness under realistic distribution shifts, including unseen disease categories and changes in radiotracers.
The first external dataset consists of tumor segmentation tasks collected from an independent clinical center, where the tumor categories correspond to previously unseen cancer types that are not present in the training data. This setting evaluates the ability of SegAnyPET to generalize to novel disease entities beyond the predefined annotation space encountered during training.
The second external dataset introduces a modality shift by incorporating PET/MRI scans. While PET/MRI offers clinical advantages by reducing ionizing radiation compared to PET/CT, it presents challenges for segmentation models due to the distinct underlying physics of attenuation correction. Unlike PET/CT, which utilizes CT-derived attenuation maps for precise quantification, PET/MRI relies on estimated tissue maps that often exhibit variances in bone and lung density representation. By evaluating SegAnyPET on this dataset, we aim to test its robustness against the domain gap inherent in different imaging architectures and reconstruction protocols.
The third external dataset is derived from PSMA-PET imaging, which employs a radiotracer that is fundamentally different from the commonly used FDG tracer in routine PET studies. PSMA-PET is a prostate-specific imaging modality that targets prostate-specific membrane antigen, enabling highly specific visualization of prostate cancer lesions. The distinct uptake patterns and image characteristics of PSMA-PET introduce an additional layer of domain shift, providing a stringent test of cross-tracer generalization.
These external evaluations aim to examine whether SegAnyPET can maintain reliable segmentation performance on previously unseen tasks, thereby validating its design as a general-purpose promptable PET segmentation model capable of adapting to diverse clinical scenarios without task-specific retraining.

\begin{figure*}[htbp]
    \centering
    \includegraphics[width=\linewidth]{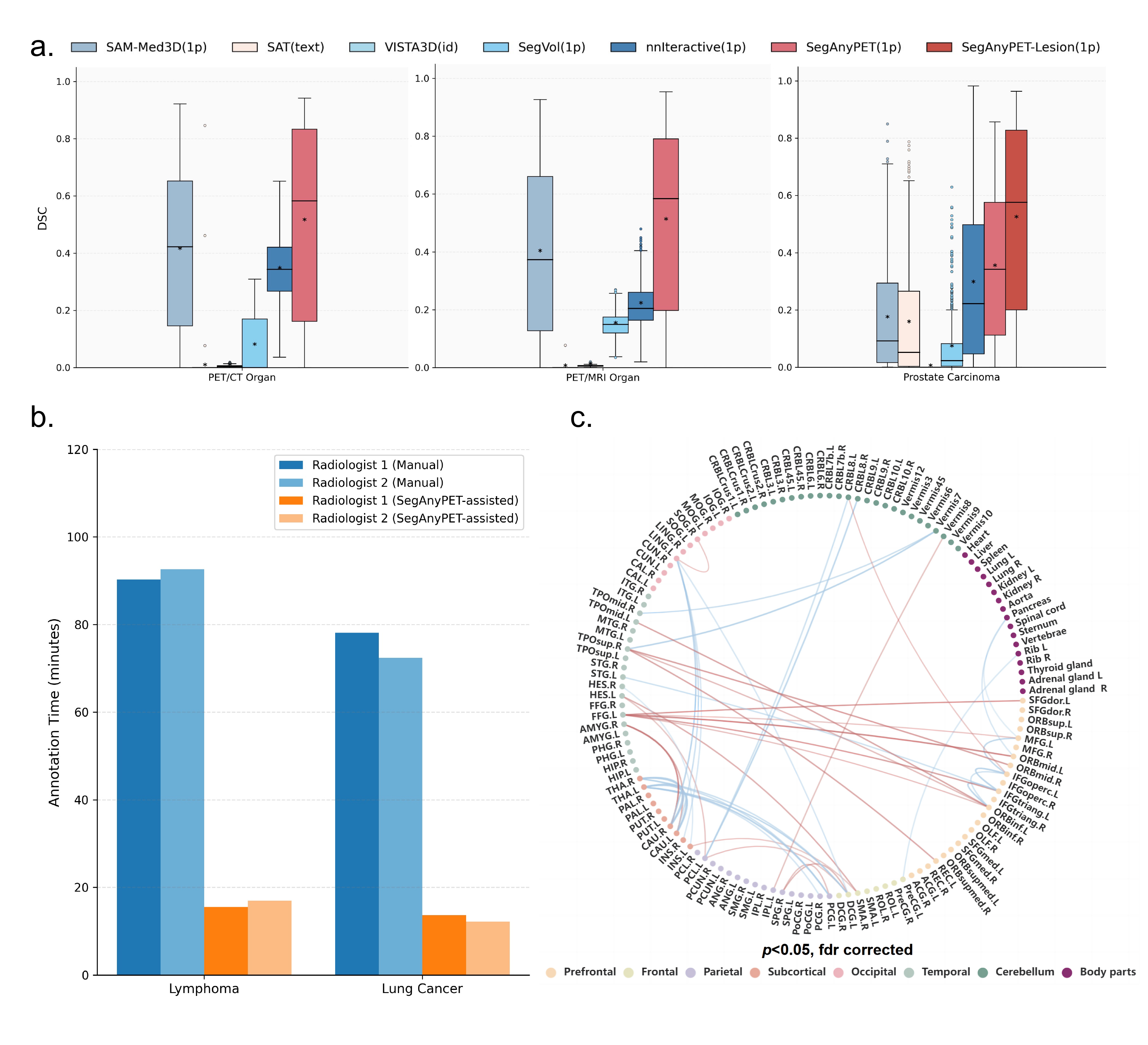}  
    \caption{\textbf{Figure 4: Comprehensive evaluation of generalization capability in external cohorts and clinical utility in downstreeam applications.} 
    (a) Quantitative comparison of model generalization performance on out-of-distribution datasets with unseen targer organs, modality variations, and novel radiotracer uptake patterns.
    (b) Evaluation of clinical utility in terms of annotation efficiency. In both lymphoma and lung cancer scenarios, the SegAnyPET-assisted interactive workflow significantly reduces the manual burden compared to conventional manual delineation.
    (c) Whole-body metabolic covariance network built from SegAnyPET segmentation results. The network verifies that SegAnyPET outputs have high biological fidelity, enabling robust downstream metabolic network analysis for systemic disease clinical research.}
    \label{fig:evaluation}
\end{figure*}

Figure \ref{fig:evaluation}(a) presents quantitative comparisons across the three external validation datasets.
As demonstrated by the quantitative metrics and the visualizations, SegAnyPET shows robust generalization capability despite significant distribution shifts, including unseen disease entities, modality variations, and novel radiotracer uptake patterns, effectively validating that SegAnyPET has encoded domain-agnostic metabolic representations and demonstrating a robust zero-shot capacity to navigate the inherent complexities of diverse real-world clinical workflows.

\subsection*{Clinical utility of SegAnyPET in downstream applications}

Beyond improvements in quantitative segmentation metrics, the practical value of SegAnyPET lies in its ability to enhance clinical workflows and support downstream oncological applications. 
In oncological practice, accurate tumor segmentation in routine practice remains a labor-intensive and subjective process, forming a critical bottleneck for clinical tasks like disease staging. However, this process remains largely dependent on manual delineation, which is time-consuming and prone to inter-observer variability.
In whole-body PET imaging, this burden is further amplified by complex lesion distributions and heterogeneous tracer uptake.

To evaluate the practical clinical utility of SegAnyPET, we conduct a human annotation study focusing on two representative oncological scenarios: lymphoma and lung cancer. We compare the time cost of two annotation pipelines: (1) a conventional clinical workflow, in which two experienced nuclear medicine physicians independently perform full manual delineation, and (2) an interactive segmentation workflow assisted by SegAnyPET, where physicians iteratively provide point-based prompts until satisfactory segmentation results are obtained.
As shown in Figure~\ref{fig:evaluation}(b), the SegAnyPET-assisted workflow leads to a substantial reduction in annotation time, with time savings of 82.37\% and 82.95\% for the two experts, respectively. These results demonstrate that prompt-based interaction can significantly alleviate the manual burden of PET tumor delineation while preserving clinical control over the final segmentation outcome.

To further evaluate the effectiveness of the proposed SegAnyPET model in practical downstream tasks, we apply it to whole-body metabolic network analysis.
The comprehensive assessment of complex systemic diseases often relies on understanding sophisticated metabolic interactions across multiple organs. The construction and analysis of systemic organ networks represent an increasingly important application scenario in clinical research \cite{sun2022identifying,ruan2025graph}.
By accurately and consistently delineating diverse regions of interest (ROIs), SegAnyPET facilitates the robust automated construction of these metabolic covariance networks. As visualized in Figure~\ref{fig:evaluation}(c), the resulting network representations effectively capture holistic inter-organ connectivity patterns, proving that the model's outputs possess the high biological fidelity required to support reliable functional quantification in systemic clinical research.
Crucially, a distinctive advantage of employing SegAnyPET for such analysis is its universal and scalable ability. In conventional deep learning paradigms, incorporating a newly targeted organ or structure into an existing metabolic network would necessitate a cycle of data collection, manual annotation, and model retraining. In contrast, SegAnyPET's prompt-driven architecture allows researchers to seamlessly and efficiently introduce new structures into the analysis pipeline on demand, simply through intuitive interaction. This flexibility provides a highly scalable technical foundation, significantly accelerating exploratory systems biology research and the comprehensive clinical assessment of systemic diseases.

\begin{figure*}[t]
    \centering
    \includegraphics[width=\linewidth]{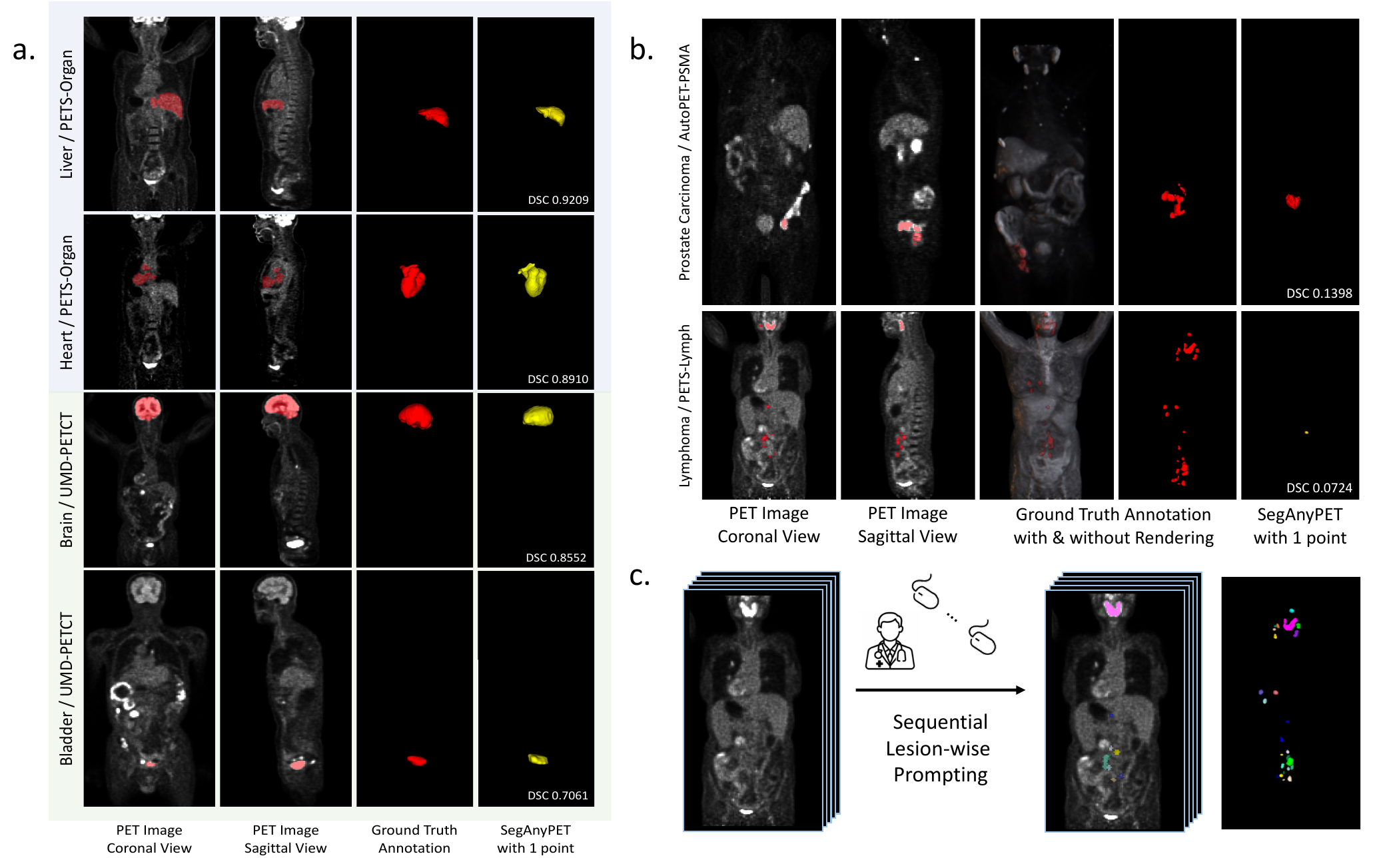}  
    \caption{\textbf{Figure 5: Comparison of point-prompt interaction efficacy between continuous organs and distributed pathological lesions.} 
    (a) Segmentation of continuous holistic organs. For such single-entity structures, precise and comprehensive segmentation can be efficiently achieved using only one point prompt. (b) Segmentation of disseminated lesion like lymphoma. In contrast to organs, systemic oncological findings frequently present as multiple, spatially discrete lesions across the whole body. This example illustrates the practical constraints of the point-based paradigm, demonstrating that a limited number of point prompts are insufficient to simultaneously capture the entire distributed tumor burden. (c) Practical interactive workflow for disseminated lesions. Since a single prompt cannot capture all spatially separated tumor sites, each lesion is individually prompted and segmented in a sequential lesion-wise manner. }
    \label{fig:visualization}
\end{figure*}

    \section*{Discussion}
        In this work, we present SegAnyPET, a foundation model for universal volumetric PET segmentation. 
While recent years have witnessed rapid progress in segmentation foundation models for natural images and structural medical images, PET imaging has remained underrepresented, not merely due to data scarcity but because of its intrinsic divergence from structural modalities, rendering direct transfer of anatomy-centric foundation models ineffective, particularly under distribution shifts.
The consistently inferior performance of existing general-purpose medical foundation models on PET tasks observed in our evaluations highlights a critical limitation of existing foundation models and underscores the necessity of PET-specific representation learning.

By leveraging large-scale, heterogeneous whole-body PET imaging, SegAnyPET captures domain-robust metabolic representations that generalize across scanners, institutions, disease entities, imaging protocols, and radiotracers, enabling reliable zero-shot segmentation in settings that remain inaccessible to conventional task-specific or adaptation-based approaches.
Besides, the prompt-based design of SegAnyPET challenges the long-standing paradigm of task-specific PET segmentation. Traditional deep learning models are intrinsically constrained by predefined annotation taxonomies and require repeated cycles of annotation and retraining to accommodate new organs, lesions, or disease patterns, a limitation that is particularly restrictive in whole-body PET imaging where clinically relevant findings are diverse, rare, and often unpredictable.
In contrast, the promptable design of SegAnyPET enables user-guided adaptation at inference time, allowing the model to handle targets outside the original training taxonomy without retraining. This capability is particularly important in whole-body PET interpretation, where lesion appearance, anatomical context, and clinical questions vary substantially across patients.
Beyond quantitative performance, our evaluations in downstream applications demonstrate the practical clinical value of SegAnyPET.
By substantially reducing annotation overhead while preserving essential physician oversight, SegAnyPET acts as a flexible tool that enhances both the speed and reproducibility of clinical annotations.
Furthermore, this practical utility translates directly to complex downstream clinical research, such as whole-body metabolic network analysis. The interactive architecture enables the seamless, on-demand extraction of novel regions of interest. This highly scalable adaptability ensures that the model can readily accommodate the diverse and often unpredictable nature of systemic diseases, ultimately providing a robust technical foundation to accelerate exploratory systems biology and comprehensive clinical assessments.

Several limitations merit consideration. While PETWB-Seg11K substantially expands the scale and diversity of PET segmentation data, certain rare diseases, radiotracers, and anatomical regions remain underrepresented. 
Furthermore, although our model demonstrates significant improvements over existing approaches in lesion segmentation tasks, the quantitative metrics indicate that there is still substantial room for enhancement.
In clinical whole-body PET imaging, oncological findings frequently present as multiple spatially discrete lesions (see Figure~\ref{fig:visualization}). Consequently, the current point-prompt-based interaction paradigm faces practical constraints to achieve comprehensive segmentation of all distributed lesions using merely one or a few discrete point clicks.
This limitation underscores the need for more efficient interaction strategy.
Recent advances in segmentation foundation models have explored text-based prompting as an alternative or complementary interaction paradigm, which could be applied to specify targets through semantic descriptions to identify all distributed lesions simultaneously.
Although existing text-prompt-driven models show limited effectiveness on PET data in our evaluation likely due to the domain gap between functional PET imaging and the predominantly structural modalities used during their training, multimodal vision-language PET foundation model remains a promising future direction \cite{zhao2025foundation}.
Integrating PET images with textual information, such as key findings extracted from radiology reports or clinical indications, may enable more intuitive and efficient specification of segmentation targets.
Such a multimodal approach could potentially reduce the need for explicit user interaction by directly generating clinically relevant segmentations from semantic cues, further streamlining the clinical workflow. We believe that advancing PET-specific multimodal foundation models, jointly leveraging imaging and language information, represents an important step toward fully automated yet clinically controllable PET analysis.

Overall, this study establishes SegAnyPET as the first and foundational step toward universal promptable PET segmentation, demonstrating that large-scale PET-specific foundation modeling can achieve strong performance, robust generalization, and tangible clinical benefits. We believe this work provides a scalable framework for advancing AI-assisted PET analysis and lays the groundwork for scalable, generalizable, and clinically trustworthy AI solutions in functional medical imaging.

%    \end{refsegment}

%\printbibliography[segment=1, heading=bibliography, title={Main References}]

    %\bibliographystyle{ieeetr}

    %\defaultbibliography{bibliography}
    %\defaultbibliographystyle{ieeetr}

    %\bibliography{bibliography.bib}
\newpage
 %   \appendix

   %  \begin{refsegment}
         
    \section*{Methods} \label{methods}

\subsection*{Dataset construction}

In this study, we construct PETWB-Seg11K, a multi-center, multi-tracer whole-Body PET dataset collected from two open-source datasets including AutoPET \cite{AutoPET} and UDPET \cite{xue2025udpet}, and two private cohorts entitled PETS-A and PETS-B.
A total of 11,041 cases were collected from multiple clinical centers around the world, which denotes the largest and most comprehensive collection of 3D PET imaging to the best of our knowledge.
For private sources, DICOM images were queried from the clinical data storage system, converted to nifti files, and de-identified before conducting any computational analysis.
We group these datasets into development set and validation set for training and testing of the model, where the details are shown in Table \ref{Table_Dataset}.

\begin{table*}
	\centering
\captionsetup{labelformat=empty, labelsep=none}
\setlength\tabcolsep{5pt}
\renewcommand\arraystretch{1.5}
\begin{tabular}{c|c|c|c|ccc}
		\hline
Cohort ID & Dataset & Modality & Segmentation Task & Num. Scans & Num. Masks \\ \hline
C1 & AutoPET-FDG & FDG-PET & Tumor Lesion &  1,014 & 9,696 \\
C2 & UDPET & FDG-PET & Multi-Organ & 1,371 & 6,855 \\
C3 & PETS-A & FDG-PET & Multi-Organ & 5,731 & 28,655 \\
C4 & PETS-B & FDG-PET & Multi-Organ & 2,925 & 14,625 \\ \hline
Dev & \multicolumn{3}{c}{PETWB-Seg11K (Model Development)} & 11,041 & 59,831 \\ \hline \hline 
C3 & PETS-Organ & FDG-PET & Multi-Organ &  100 & 1,100 \\
C3 & PETS-Lymph & FDG-PET & Lymphoma & 148 & 414  \\ 
C3 & PETS-LC & FDG-PET & Lung Cancer & 638 & 8,314   \\ \hline
Int & \multicolumn{3}{c}{PETWB-Int (Internal Validation) } & 886 & 9828 \\ \hline
C5 & UMD-PETCT & FDG-PET & Multi-Organ & 490 & 6,370 \\
C6 & UMD-PETMR & FDG-PET* & Multi-Organ &  464 & 6,032 \\  
C7 & AutoPET-PSMA & PSMA-PET & Prostate Carcinoma &  597 & 22,177 \\ \hline
Ext & \multicolumn{3}{c}{PETWB-Ext (External Validation) } & 1,551 & 34,579 \\ \hline \hline 
\multicolumn{4}{c}{}\\
\end{tabular}
\caption{\textbf{Table 1}: Information summary of subsets PETWB-Seg11K dataset involved in the development and evaluation of SegAnyPET. * denotes PET/MRI-derived PET imaging other than PET/CT-derived PET imaging}
\label{Table_Dataset}
\end{table*}

The Development Set contains 11,041 whole-body PET scans and serves as the primary source for model training. Below we provide details of the datasets. \textbf{C1: AutoPET-FDG dataset.} The AutoPET dataset \cite{AutoPET} contains 1,014 whole-body FDG-PET/CT scans acquired from patients with histologically confirmed melanoma, lymphoma, or lung cancer, as well as control subjects without detectable FDG-avid lesions.
Among the dataset, 513 cases have no lesions, whereas 188, 168, and 145 cases correspond to melanoma, lung cancer, and lymphoma, respectively.
The dataset was collected from University Hospital Tübingen and Germany University Hospital of the LMU in Munich.
\textbf{C2: UDPET dataset.} The Ultra-Low Dose PET Imaging (UDPET) dataset consists of 1,371 subjects who underwent paired full-dose and ultra-low-dose FDG-PET acquisitions. Scans were obtained using two state-of-the-art long-axial field-of-view scanners: Siemens Biograph Vision Quadra (n = 387) and United Imaging uEXPLORER (n = 1,060).
The Quadra data was collected from University of Bern and uExplorer data was collected from Ruijin Hospital, Shanghai Jiao Tong University School of Medicine. 
In this work, we utilize full-dose PET images for model development due to their superior signal-to-noise ratio and more reliable anatomical visualization for segmentation.
\textbf{C3: PETS-A dataset.} The PETS-A cohort includes 5,731 whole-body 18F-FDG PET/CT scans acquired at The Second Affiliated Hospital Zhejiang University using a Siemens Biograph 64 PET/CT system in 3D mode. All patients fasted for at least 6 hours prior to imaging and had blood glucose levels below 200 mg/dL before examination. PET scans were performed approximately 60 minutes after intravenous administration of 18F-FDG at a dose of 3.70 MBq/kg. Low-dose CT images were used for attenuation correction for all PET acquisitions.
\textbf{C4: PETS-B dataset.} The PETS-B cohort consists of 2,925 whole-body 18F-FDG PET/CT scans collected at Hangzhou Universal Medical Imaging Diagnostic Center using a GE Discovery 710 PET/CT system in 3D mode. Subjects fasted for at least 6 hours before PET acquisition, followed by intravenous injection of 18F-FDG at 3.7 MBq/kg. Images were acquired after a 40-minute uptake period with subjects’ eyes open. PET volumes were reconstructed using ordered subset expectation maximization and attenuation correction was performed using vendor-provided CT-based correction.

In addition to the development set, we incorporate both internal and external validation subsets to comprehensively evaluate the generalization capability of SegAnyPET across scanners, institutions, and clinical indications. The internal validation cohorts were derived from the same clinical center (C3) but include independent subjects with expert-verified annotations to ensure unbiased evaluation. Specifically, PETS-Organ provides annotations for 11 major organs including liver, left kidney, right kidney, heart, spleen, aorta, left lung lower lobe, right lung lower lobe, left lung upper lobe, right lung upper lobe, and right lung middle lobe for enabling fine-grained assessment of multi-organ segmentation quality.
For lesion segmentation, PETS-Lymph and PETS-LC consist of patients diagnosed with lymphoma and lung cancer, respectively, and contains expert delineations of all tumor lesions.
The external validation cohorts were sourced from additional medical centers using different scanner modes and imaging protocols including unseen disease entities, modality variations, and novel radiotracer uptake patterns compared to the development set, enabling assessment under real-world application variability. Below we provide details of the datasets.
\textbf{C5: UMD-PETCT dataset.} The dataset contains 490 $^{18}\text{F-FDG}$ PET/CT images collected from Shanghai Universal Medical Imaging Diagnostic Center. The PET/CT imaging was performed using a Siemens Biograph 64 scanner following a 6-hour fast and confirmation of blood glucose $< 11.1 \text{ mmol/L}$. Patients received an intravenous injection of $^{18}\text{F-FDG}$ at a weight-based dose of $3.70-5.55 MBq/kg$, followed by a 60-minute uptake period. Data were acquired from the skull base to the mid-thigh, utilizing a low-dose CT ($120 \text{ kV}$, $170 \text{ mA}$) for attenuation correction and 3D PET acquisition at $2.5 \text{ min/bed}$, subsequently reconstructed via the OSEM algorithm with a $5\text{-mm}$ Gaussian post-filter.
\textbf{C6: UMD-PETMRI dataset.} The dataset contains 464 $^{18}\text{F-FDG}$ PET/MRI images collected from Shanghai Universal Medical Imaging Diagnostic Center. Simultaneous PET/MRI data were acquired on a $3.0\text{T}$ Siemens Biograph mMR system covering the head to the mid-thigh. Subjects received $^{18}\text{F-FDG}$ at $3.7 \text{ MBq/kg}$ and rested for 45 minutes prior to a $3-4 \text{ min/bed}$ 3D PET scan. Attenuation correction was performed using a Dixon-based four-segment $\mu\text{-map}$ derived from a volumetric interpolated breath-hold examination, ensuring optimal spatiotemporal alignment between the functional PET and anatomical MRI sequences.
The UMD dataset contains voxel-wise annotations for 13 organs, including the liver, left kidney, right kidney, brain, heart, spleen, aorta, lung, colon, urinary bladder, pancreas, esophagus, and stomach. Notably, the inclusion of additional organs unseen during the training of SegAnyPET enables an effective evaluation of its generalization performance.
\textbf{C7: AutoPET-PSMA dataset.} The dataset contains 597 whole-body PET imaging of male patients with prostate carcinoma, comprising 537 studies with PSMA-avid tumor lesions and 60 studies without. The studies were acquired from the LMU Hospital in Munich using three different PET/CT scanners including Siemens Biograph 64-4R TruePoint, Siemens Biograph mCT Flow 20, and GE Discovery 690.

\begin{figure*}[htbp]
    \centering
    \includegraphics[width=\linewidth]{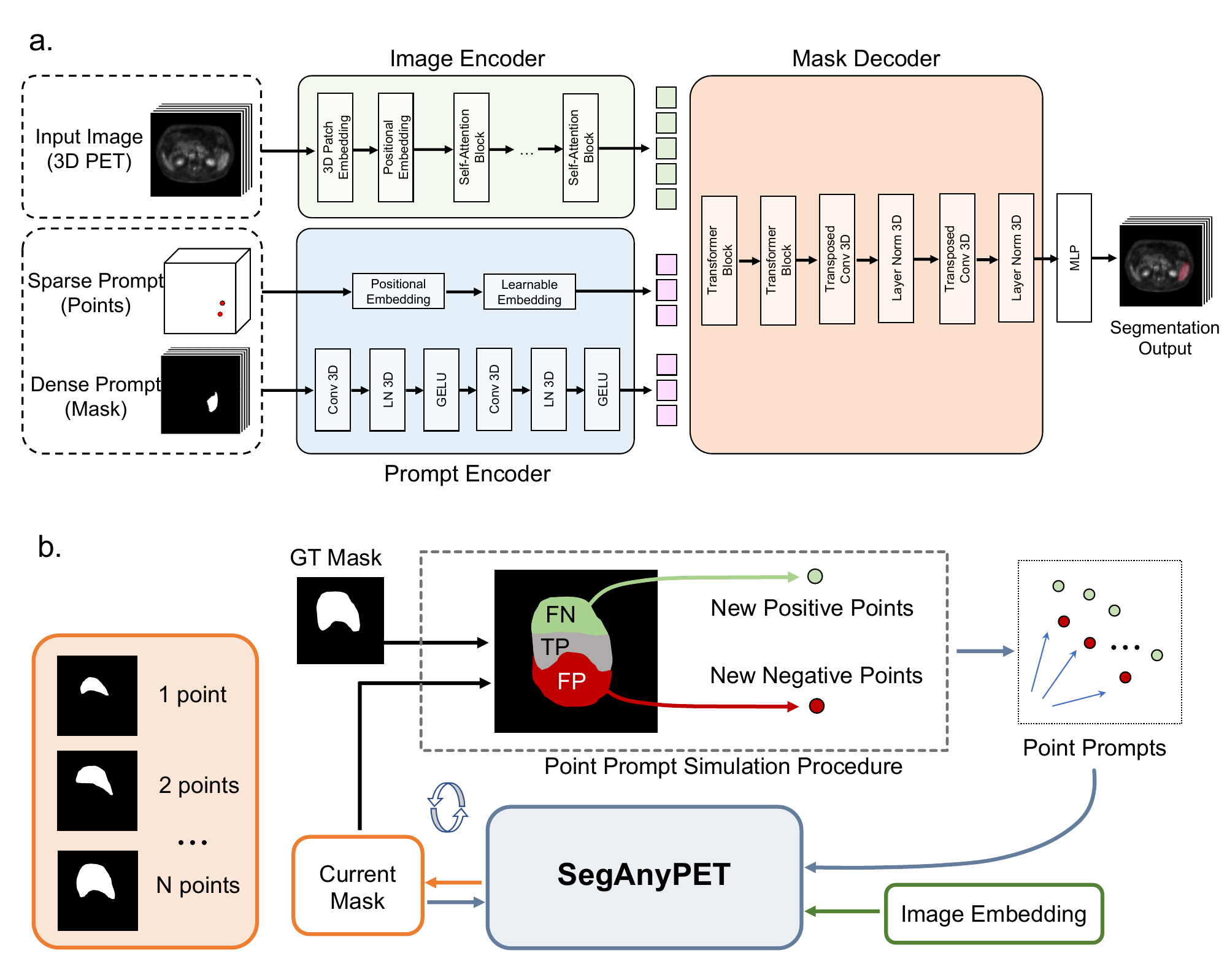}  
    \caption{\textbf{Figure 6: Network architectural and promptable segmentation pipeline of SegAnyPET for volumetric PET segmentation.} 
    (a) SegAnyPET employs 3D transformer-based components to directly capture spatial correlations within volumetric data. The pipeline integrates 3D image feature extraction with absolute positional encoding, a prompt encoder supporting both sparse (points) and dense (masks) inputs, and a mask decoder process utilizing 3D transformer blocks and transposed convolutions for end-to-end volumetric segmentation. (b) The iterative evaluation loop simulates an interactive segmentation process by automatically generating clicks based on the discrepancy between the current mask prediction and the ground truth. These accumulating points, combined with the previous mask and image embeddings, are fed back into SegAnyPET to progressively refine the prediction.}
    \label{fig:architecture}
\end{figure*}

\subsection*{Model architecture and promptable segmentation pipeline}

As shown in Figure \ref{fig:architecture}(a), the architecture of SegAnyPET follows the foundational design principles of SAM, where the original 2D transformer-based components are extended to a fully 3D formulation suitable for volumetric medical imaging.
Instead of compromised solutions like 2D framework with 3D adapters or slice-by-slice aggregation, this design can directly capture 3D spatial correlations in volumetric data. 
The model consists of three core components: an Image Encoder, a Prompt Encoder, and a Mask Decoder to enable end-to-end processing of volumetric PET images with dimensions $X \in R^{H \times W \times D}$.
The Image Encoder $\mathcal{E}_{I}$ takes raw PET volumes as input and produces high-dimensional feature embeddings by integrating 3D absolute positional encoding to preserve spatial location information.
To flexibly condition the segmentation process, the Prompt Encoder $\mathcal{E}_{P}$ supports both sparse and dense 3D prompts and converts then into compatible embeddings, incorporating learnable parameters that adapt to the characteristics of PET images and enhance the alignment between prompt information and image features. Sparse prompts (i.e. points) are encoded via 3D positional embeddings, while dense prompts (i.e. coarse masks) 
are processed through a dedicated pathway of 3D convolutional layers, 3D layer normalization, and GELU activations to align seamlessly with the image latent space.
The Mask Decoder $\mathcal{D}$ then fuses the image and prompt embeddings through 3D transformer blocks, followed by progressive upsampling via 3D transposed convolutions and a prediction head, producing volumetric segmentation masks.

To simulate and rigorously evaluate an interactive segmentation process, SegAnyPET employs an iterative evaluation loop, as shown in Figure \ref{fig:architecture}(b). During each iterative step $t$, the system automatically generate new point prompt by comparing the current predicted mask $\hat{Y}^{(t)}$ against the ground truth mask $Y$. The discrepancies between the two masks directly drive the generation of new prompts. Specifically, new positive points are strategically sampled from false-negative (FN) regions, mathematically formulated as $p^{(t)}_{pos} \in Y \setminus \hat{Y}^{(t)}$, while new negative points are sampled from false-positive (FP) regions, formulated as $p^{(t)}_{neg} \in \hat{Y}^{(t)} \setminus Y$. These continuously accumulating prompt points, along with the previously generated mask and the static image embeddings, are fed back into SegAnyPET for the subsequent prediction round, defined as:
\begin{equation}
    \hat{Y}^{(t+1)} = \mathcal{D}(\mathcal{E}_{I}(X), \mathcal{E}_P(p^{(t)}, \hat{Y}^{(t)}))
\end{equation}
This continuous iterative feedback loop allows the model to progressively refine its predictions, effectively driving the segmentation output to converge rapidly toward the anatomical ground truth.

\begin{figure*}[tbp]
    \centering
    \includegraphics[width=\linewidth]{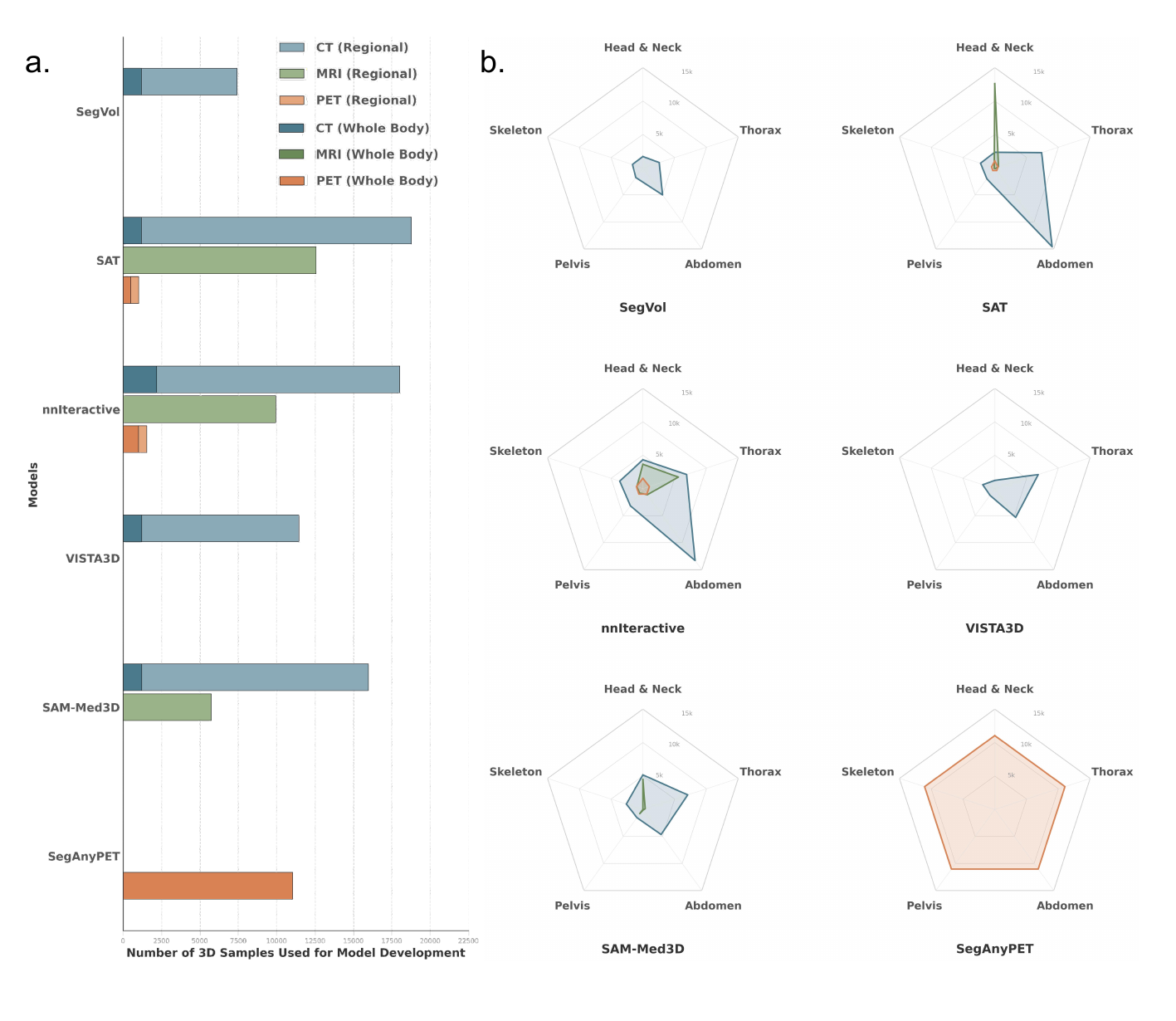} 
    \caption{\textbf{Figure 7: Training data composition of representative 3D medical segmentation foundation models.} 
    (a) Number of 3D training samples across imaging modalities (CT, MRI, and PET), with whole-body and regional datasets shown separately. This modality disparity underscores the fundamental limitation of current models, which are heavily biased toward structural priors and struggle to represent complex metabolic uptake patterns.
    (b) Anatomical coverage of the training data across major body regions. Most existing models are trained on structural CT or MRI data with region-specific coverage, whereas SegAnyPET is trained on large-scale whole-body PET data spanning all major anatomical regions. By bridging this critical data scarcity with comprehensive whole-body PET volumes, SegAnyPET establishes a robust and domain-specific feature space tailored for functional PET image analysis.}
    \label{fig:dataused}
\end{figure*}

\subsection*{Training protocol and experimental setting}

Our method is implemented in Python with PyTorch and trained on 8 NVIDIA Tesla A100 GPUs, each with 80GB memory. 
We use the AdamW optimizer with an initial learning rate of 8e-5 for the image encoder, while the prompt encoder and mask decoder are optimized with a 10x smaller learning rate, and a weight decay factor of 0.1 is applied to all parameters.
The training of SegAnyPET was performed for a total of 500 epochs on the constructed dataset with a volumetric input patch size of 128$\times$128$\times$128.
We further derived a specialized variant SegAnyPET-Lesion by fine-tuning on lesion-centric training data for 200 additional epochs.
In the distributed training setup, each GPU is configured with a micro-batch size of 12, leading to a global batch size of 96 across 8 GPUs.
To handle the learning rate schedule, we employed the MultiStepLR scheduler, which adjusts the learning rate in predefined steps with 120 and 180 epochs, with a gamma value of 0.1, indicating that the learning rate is reduced by 10\% of its current value at each step.
The loss function is a combination of Dice loss and cross-entropy loss with sigmoid activation and squared predictions to optimize the 3D segmentation task.
For each training iteration, the number of interactive clicks for prompt guidance is randomly sampled between 1 and 20 to enhance the model's adaptability to different interaction scenarios. Additionally, data augmentation strategies including random flipping along axial, coronal, and sagittal axes and adaptive cropping/padding to the target 128$\times$128$\times$128 size are applied to improve generalization.
Mixed precision training is enabled to accelerate training and reduce memory consumption, while distributed data parallel (DDP) with NCCL backend is used for multi-GPU communication.
For inference on large volumetric data, a patch-based strategy is employed. Initially, a patch of size 128$\times$128$\times$128 is cropped around the starting point. If the predicted region reaches the patch boundary, a sliding-window approach with 50\% overlap is applied.

\subsection*{Implementation and analysis of competing methods}

To thoroughly evaluate the performance of SegAnyPET, we conducted comparation experiments against both 2D \cite{SAM,MedSAM} and 3D \cite{SAM-Med3D,zhao2025large,he2025vista3d,isensee2025nninteractive,du2024segvol} segmentation foundation models, and task-specific specialist segmentation models \cite{isensee2020nnunet,segresnet,hatamizadeh2021swin,huang2023stunet}.
Since the original SAM \cite{SAM} and MedSAM \cite{MedSAM} are designed for 2D segmentation tasks and cannot handle 3D inputs directly, a slice-by-slice procedure is conducted for the segmentation of the volume. As a result, the segmentation procedure of 2D foundation models necessitate input prompts for each 2D slice containing the target.
Consequently, employing 2D models for volumetric segmentation incurs a prompting workload that is tens of times greater, accompanied by a multi-fold increase in inference time (Details are provided in Supplementary Table \ref{tab:sp3}).
Additionally, as 2D slices fail to utilize the spatial information inherent in 3D volumetric images \cite{zhang2022bridging}, the segmentation results produced are often spatially discontinuous and require further correction.
In contrast, 3D segmentation foundation models can be directly utilized to segment the targets from input volume, instead of slice-by-slice prompting of 2D models.

Therefore, our comparative analysis primarily focuses on 3D segmentation foundation models. 
These 3D foundation models possess diverse network architectures and are trained on their distinct aggregations of open-source and private datasets.
Based on the provided data descriptions, we systematically quantified the distribution of their training data across imaging modalities and anatomical coverage, as illustrated in Figure \ref{fig:dataused}, which reveals a stark modality discrepancy.
The training corpora for most existing models are overwhelmingly dominated by structural imaging like CT and MRI, and predominantly rely on regional scans rather than whole-body volumes. 
This profound disparity in both modality and spatial coverage constitutes a fundamental limitation. Because these models are heavily biased toward localized structural priors, they inherently struggle to capture and represent the complex, systemic metabolic uptake patterns characteristic of functional imaging.
Consequently, this leads to notably poor generalization when applied to whole-body PET segmentation tasks.
In contrast, SegAnyPET specifically addresses this critical data scarcity by utilizing large-scale, whole-body PET volumes spanning all major anatomical regions for its development. By comprehensively bridging this domain gap, SegAnyPET establishes a robust and domain-specific feature space tailored for PET imaging.

Among the evaluated models, the prompt mechanisms vary significantly, encompassing architectures guided by semantic textual or target id prompts and those driven by spatial positional prompts to assign the targets for segmentation.
To ensure a rigorous and equitable comparison, we utilized the latest official and largest-scale pre-trained weights (i.e. SAT-Pro for \cite{zhao2025large} and SAM-Med3D-turbo for \cite{SAM-Med3D}) and codebases for all comparing models.
For text-driven 3D foundation models, we utilize the category names of the target organs (e.g. Liver) or lesion types (e.g. Lymphoma) as textual prompts for segmentation. For models employing positional prompts, considering that bounding boxes are often cumbersome and less intuitive in 3D volumetric space \cite{isensee2025nninteractive}, we standardize our comparative experiments by using point prompts simulated based on the ground-truth mask for interactive segmentation following the implementation of each competing method.
For specialist models, we adopted the nnUNet framework \cite{isensee2020nnunet} for data preprocessing and model training. As a self-configuring method, nnUNet automatically adapts its hyperparameter settings like image preprocessing, data augmentation, and training details based on the specific fingerprint of given dataset. This choice has been widely recognized for its robust performance across a diverse range of medical image segmentation tasks and provides a standardized pipeline that facilitates state-of-the-art performance for task-specific segmentation models \cite{isensee2024nnu}.

\subsection*{Evaluation metrics}

We quantitatively evaluate the segmentation performance from the perspective of region and boundary metrics \cite{maier2024metrics}.
For region-based evaluation, we use the Dice Similarity Coefficient (DSC), which is one of the most commonly used evaluation metric for image segmentation to measure the degree of voxel-wise overlap between the segmentation results $S$ and the ground truth $G$. Higher DSC indicate better segmentation performance. The calculation is defined as follows:
\begin{equation}
DSC(G, S) = \frac{2|G\cap S|}{|G| + |S|}
\label{DSC}
\end{equation}

In addition, we use Normalized Surface Distance (NSD), which is a boundary-based metric that measures the consistency at the boundary area of the segmentation results $S$ and the ground truth $G$, which is defined as:
\begin{equation}
NSD(G, S) = \frac{|\partial S \bigcap B_{\partial G}| + |\partial G \bigcap B_{\partial S}|}{|\partial S| + |\partial G|},
\label{eq:nsd}
\end{equation}
where $B_{\partial S} = \{x \in \mathbb{R}^3 | \exists \hat{x} \in \partial S, ||x - \hat{x}|| \leq \tau\}$ and $B_{\partial G} = \{x \in \mathbb{R}^3 | \exists \hat{x} \in \partial G, ||x - \hat{x}|| \leq \tau\}$ are the boundary areas of the segmentation results and ground truth at a tolerance $\tau$, respectively. We set $\tau$ as 1 in the experiments.

\subsection*{Data availability}
\label{dataav_methods}
Refs \cite{AutoPET,xue2025udpet} provide access to the AutoPET-FDG, AutoPET-PSMA (\url{http://autopet-iii.grand-challenge.org}) and UDPET (\url{http://udpet-challenge.github.io}) datasets used in our study, which are public by other contributors.
External validation datasets (UMD-PETCT and UMD-PETMR) will be public available at (\url{https://github.com/YichiZhang98/UMD}).
Other development and internal validation datasets (PETS-A, PETS-B, PETS-Organ, PETS-Lymph, PETS-LC) are subject to controlled access.

\subsection*{Code availability}
Our trained models and code are publicly available at \url{https://github.com/YichiZhang98/SegAnyPET} for further research.

\subsection*{Competing Interests}
The authors declare no competing interests.

   % \end{refsegment}
    %\printbibliography[segment=2, heading=bibliography, title={Methods References}, notkeyword=cited-in-main]

    \bibliographystyle{ieeetr}
    \bibliography{bibliography.bib}
    \newpage
    
        \begin{table}[htbp]
  \centering
  \captionsetup{labelformat=empty, labelsep=none}
  \resizebox{\textwidth}{!}{% 自动调整宽度以适应页面
  \begin{tabular}{lccccc}
    \toprule
    \textbf{Model} & \textbf{Liver} & \textbf{Kidney-L} & \textbf{Kidney-R} & \textbf{Heart} & \textbf{Spleen} \\
    \midrule
    nnUNet        & 0.9379 (0.9298--0.9419) & 0.9026 (0.8920--0.9155) & 0.8704 (0.8513--0.8893) & 0.9118 (0.9016--0.9180) & 0.8869 (0.8635--0.9010) \\
    SegResNet     & 0.9362 (0.9291--0.9410) & 0.9071 (0.8822--0.9188) & 0.8585 (0.8415--0.8829) & 0.9031 (0.8956--0.9137) & 0.8938 (0.8774--0.9052) \\
    STUNet        & 0.9351 (0.9265--0.9407) & 0.9031 (0.8941--0.9143) & 0.8769 (0.8315--0.8958) & 0.9176 (0.8960--0.9315) & 0.8795 (0.8592--0.9013) \\
    SwinUNETR     & 0.9351 (0.9291--0.9413) & 0.8991 (0.8788--0.9102) & 0.8461 (0.8355--0.8968) & 0.8904 (0.8852--0.9119) & 0.8857 (0.8570--0.8994) \\
    \midrule
    SegAnyPET (1p)& 0.9262 (0.9156--0.9331) & 0.8746 (0.8520--0.8877) & 0.8700 (0.8546--0.8880) & 0.8922 (0.8818--0.9002) & 0.8763 (0.8560--0.8944) \\
    SegAnyPET (3p)& 0.9386 (0.9298--0.9460) & 0.8867 (0.8662--0.9020) & 0.8875 (0.8682--0.9031) & 0.9052 (0.8967--0.9136) & 0.8908 (0.8732--0.9200) \\
    SegAnyPET (5p)& 0.9494 (0.9404--0.9558) & 0.8982 (0.8797--0.9128) & 0.8982 (0.8824--0.9150) & 0.9157 (0.9074--0.9234) & 0.9054 (0.8849--0.9209) \\
    \bottomrule
  \end{tabular}%
  }
  \vspace{5pt} 
    \caption{\textbf{Supplementary Table 1}: Quantitative comparison of DSC performance across task-specific models and SegAnyPET under different prompting conditions. The results are presented as the median Dice score along with the interquartile range (Q1--Q3).}
  \label{tab:sp1}
\end{table}

\begin{table}[htbp]
  \centering
  \captionsetup{labelformat=empty, labelsep=none}
  \resizebox{\textwidth}{!}{
  \begin{tabular}{lccccc}
    \toprule
    \textbf{Model} & \textbf{Liver} & \textbf{Kidney-L} & \textbf{Kidney-R} & \textbf{Heart} & \textbf{Spleen} \\
    \midrule
    nnUNet & 0.5076 (0.4615--0.5402) & 0.5204 (0.4804--0.5995) & 0.5257 (0.4549--0.5985) & 0.4737 (0.4280--0.5175) & 0.5231 (0.4439--0.5715) \\
    SegResNet & 0.4818 (0.4418--0.5399) & 0.5288 (0.4768--0.5871) & 0.4954 (0.4213--0.5849) & 0.4625 (0.4308--0.5137) & 0.5116 (0.4499--0.5688) \\
    STUNet & 0.4919 (0.4623--0.5325) & 0.5259 (0.4845--0.6064) & 0.5118 (0.4411--0.5872) & 0.4702 (0.4139--0.5033) & 0.5145 (0.4540--0.5573) \\
    SwinUNETR & 0.4801 (0.4157--0.5168) & 0.4960 (0.4249--0.5649) & 0.4641 (0.3943--0.5530) & 0.4597 (0.4137--0.4865) & 0.4781 (0.4264--0.5214) \\
    \midrule
    SegAnyPET (1p)& 0.4982 (0.4550--0.5364) & 0.5123 (0.4681--0.5754) & 0.5170 (0.4512--0.5827) & 0.4689 (0.4257--0.5094) & 0.5096 (0.4475--0.5608) \\
    SegAnyPET (3p)& 0.5225 (0.4758--0.5614) & 0.5394 (0.4956--0.6028) & 0.5441 (0.4785--0.6079) & 0.4873 (0.4421--0.5286) & 0.5378 (0.4689--0.5884) \\
    SegAnyPET (5p)& 0.5486 (0.5032--0.5897) & 0.5631 (0.5184--0.6237) & 0.5674 (0.5023--0.6315) & 0.5068 (0.4617--0.5481) & 0.5659 (0.4974--0.6152) \\
    \bottomrule
  \end{tabular}
  }
  \vspace{5pt} 
  \caption{\textbf{Supplementary Table 2}: Quantitative comparison of NSD performance across task-specific models and SegAnyPET under different prompting conditions. The results are presented as the median NSD score along with the interquartile range (Q1--Q3). }
  \label{tab:sp2}
\end{table}

\begin{table}[htbp]
  \centering
  \captionsetup{labelformat=empty, labelsep=none}
  \resizebox{\textwidth}{!}{
  \begin{tabular}{cccccccccc}
    \toprule
\multirow{2}{*}{\textbf{Methods}} & \multirow{2}{*}{\textbf{Prompt}} & \multicolumn{6}{c}{\textbf{DSC Performance of Each Target Organ}} & \textbf{Inference} \\
\cline{3-8} 
&& \textbf{Liver} & \textbf{Kidney-L} & \textbf{Kidney-R} & \textbf{Brain} & \textbf{Heart} & \textbf{Spleen} &  \textbf{Time (s)} \\ \midrule
SAM  & 3N points & 0.2607 (0.2277--0.2976) & 0.2751 (0.2391--0.3136) & 0.2619 (0.2218--0.2948) & 0.1764 (0.1545--0.2017) & 0.2238 (0.1835--0.2611) & 0.3132 (0.2685--0.3579) & 19.6 \\
SAM  & 5N points & 0.3673 (0.3257--0.4014) & 0.3450 (0.3067--0.3834) & 0.3518 (0.3082--0.3937) & 0.2249 (0.2014--0.2542) & 0.3119 (0.2754--0.3522) & 0.3944 (0.3490--0.4441) & 32.1 \\
MedSAM  & 5N points & 0.3488 (0.3211--0.3799) & 0.3244 (0.2792--0.3667) & 0.3349 (0.2791--0.3836) & 0.2364 (0.2225--0.2489) & 0.2286 (0.2103--0.2502) & 0.3212 (0.2760--0.3757) & 22.8 \\ \midrule
nnIteractive & 1 point & 0.2955 (0.0040--0.7949) & 0.5970 (0.0065--0.7687) & 0.4816 (0.0104--0.7612) & 0.9027 (0.8871--0.9151) & 0.0649 (0.0088--0.5149) & 0.5020 (0.0279--0.7914) & 7.4 \\
SAM-Med3D & 1 point & 0.7020 (0.6346--0.7504) & 0.5894 (0.4877--0.6751) & 0.6083 (0.4376--0.7040) & 0.8101 (0.6313--0.8582) & 0.4661 (0.3779--0.5584) & 0.6136 (0.5192--0.6702) & 1.8 \\
SegAnyPET & 1 point & 0.8931 (0.8729--0.9083) & 0.8078 (0.7668--0.8363) & 0.8180 (0.7750--0.8507) & 0.8978 (0.8696--0.9145) & 0.6267 (0.5608--0.6796) & 0.8422 (0.8062--0.8666) & 1.8 \\
    \bottomrule
  \end{tabular}
  }
  \vspace{5pt} 
  \caption{\textbf{Supplementary Table 3}: Performance comparison with best-performing 2D and 3D foundation models for organ segmentation in UMD-PETCT with corresponding prompting efforts and inference time per case. N denotes the number of slices containing the target organ ranging from around 15 to 30 in our task. The results are presented as the median NSD score along with the interquartile range (Q1--Q3).}
  \label{tab:sp3}
\end{table}

\begin{table}[htbp]
  \centering
  \captionsetup{labelformat=empty, labelsep=none}
  \resizebox{\textwidth}{!}{% 自动调整宽度以适应页面
  \begin{tabular}{l|ccccc|cc}
    \toprule
    \textbf{Model} & \textbf{SAM-Med3D} & \textbf{SAT} & \textbf{VISTA3D} & \textbf{SegVol} & \textbf{nnIteractive} & \textbf{SegAnyPET(1p)} & \textbf{SegAnyPET-Lesion} \\ \midrule
    \textbf{Prompt} & 1 point & text & target id & 1 point & 1 point & 1 point & 1 point \\
    \midrule
    \multicolumn{8}{c}{\textbf{PETS-Organ for Multi-Organ Segmentation}} \\
    \midrule
    Liver & 0.6455 (0.3114--0.7590) & 0.0000 (0.0000--0.0000) & 0.0512 (0.0309--0.0848) & 0.4398 (0.0629--0.6184) & 0.7021 (0.0162--0.8533) & \textbf{0.9262 (0.9156--0.9331)} & --- \\
    Kidney-L & 0.6635 (0.5631--0.7200) & 0.0000 (0.0000--0.0000) & 0.0000 (0.0000--0.0000) & 0.1521 (0.0992--0.2639) & 0.7897 (0.5655--0.8305) & \textbf{0.8746 (0.8520--0.8877)} & --- \\
    Kidney-R & 0.5812 (0.4768--0.6290) & 0.0000 (0.0000--0.0000) & 0.0000 (0.0000--0.0000) & 0.1274 (0.0793--0.2432) & 0.7493 (0.3269--0.8165) & \textbf{0.8700 (0.8546--0.8880)} & --- \\
    Heart & 0.6112 (0.4622--0.6846) & 0.0000 (0.0000--0.0000) & 0.0000 (0.0000--0.0000) & 0.3459 (0.1143--0.6201) & 0.1518 (0.0239--0.6027) & \textbf{0.8922 (0.8818--0.9002)} & --- \\
    Spleen & 0.4190 (0.3248--0.5260) & 0.0000 (0.0000--0.0000) & 0.0000 (0.0000--0.0000) & 0.2265 (0.0861--0.4653) & 0.8150 (0.6701--0.8568) & \textbf{0.8763 (0.8560--0.8944)} & --- \\
    Aorta & 0.1579 (0.1161--0.2401) & 0.0000 (0.0000--0.0000) & 0.0000 (0.0000--0.0000) & 0.1680 (0.0782--0.2436) & 0.2076 (0.0714--0.2981) & \textbf{0.8150 (0.7881--0.8397)} & --- \\
    Lung Lobe & 0.3985 (0.2655--0.4876) & 0.0000 (0.0000--0.0000) & 0.0000 (0.0000--0.0000) & 0.0708 (0.0359--0.0955) & 0.2419 (0.1565--0.3346) & \textbf{0.7207 (0.6568--0.7577)} & --- \\
    \midrule
    \multicolumn{8}{c}{\textbf{PETS-Lymph for Lymphoma Lesion Segmentation}} \\
    \midrule
    Lymphoma & 0.0154 (0.0034--0.0640) & 0.0076 (0.0000--0.0569) & 0.0000 (0.0000--0.0000) & 0.0075 (0.0003--0.0337) & 0.0286 (0.0013--0.3643) & 0.0892 (0.0315--0.2076) & \textbf{0.0932 (0.0236--0.2711)} \\ \midrule
    \multicolumn{8}{c}{\textbf{PETS-LC for Lung Cancer Lesion Segmentation}} \\ \midrule
    Lung Cancer & 0.0350 (0.0130--0.0860) & 0.0000 (0.0000--0.0000) & 0.0000 (0.0000--0.0000) & 0.0634 (0.0220--0.1205) & 0.0638 (0.0195--0.1292) & 0.1085 (0.0234--0.3071) & \textbf{0.1528 (0.0392--0.3760)} \\
    \midrule
    \multicolumn{8}{c}{\textbf{UMD-PETCT for Multi-Organ Segmentation}} \\ \midrule
    Aorta & 0.1608 (0.1236--0.2009) & 0.0000 (0.0000--0.0000) & 0.0000 (0.0000--0.0000) & 0.0000 (0.0000--0.0973) & 0.0223 (0.0009--0.0842) & \textbf{0.2134 (0.1884--0.2866)} & --- \\
    Brain & 0.8101 (0.6313--0.8582) & 0.0000 (0.0000--0.0000) & 0.0000 (0.0000--0.0000) & 0.0000 (0.0000--0.2007) & \textbf{0.9027 (0.8871--0.9151)} & 0.8978 (0.8696--0.9145) & --- \\
    Colon & 0.1239 (0.0839--0.1740) & 0.0000 (0.0000--0.0000) & 0.0000 (0.0000--0.0000) & 0.0000 (0.0000--0.0336) & 0.0358 (0.0090--0.1031) & \textbf{0.1519 (0.0916--0.2182)} & --- \\
    Esophagus & 0.0564 (0.0417--0.0778) & 0.0000 (0.0000--0.0000) & 0.0000 (0.0000--0.0000) & 0.0000 (0.0000--0.0381) & 0.0467 (0.0316--0.1014) & \textbf{0.1441 (0.0992--0.1956)} & --- \\
    Heart & 0.4661 (0.3779--0.5584) & 0.0000 (0.0000--0.0000) & 0.0000 (0.0000--0.0000) & 0.0000 (0.0000--0.1458) & 0.0649 (0.0088--0.5149) & \textbf{0.6267 (0.5608--0.6796)} & --- \\
    Kidney-L & 0.5894 (0.4877--0.6751) & 0.0000 (0.0000--0.0000) & 0.0000 (0.0000--0.0000) & 0.0000 (0.0000--0.0753) & 0.5970 (0.0065--0.7687) & \textbf{0.8078 (0.7668--0.8363)} & --- \\
    Kidney-R & 0.6083 (0.4376--0.7040) & 0.0000 (0.0000--0.0000) & 0.0000 (0.0000--0.0000) & 0.0000 (0.0000--0.0811) & 0.4816 (0.0104--0.7612) & \textbf{0.8180 (0.7750--0.8507)} & --- \\
    Liver & 0.7020 (0.6346--0.7504) & 0.0000 (0.0000--0.0000) & 0.0583 (0.0317--0.0901) & 0.0000 (0.0000--0.3225) & 0.2955 (0.0040--0.7949) & \textbf{0.8931 (0.8729--0.9083)} & --- \\
    Lung & \textbf{0.5603 (0.4376--0.6479)} & 0.0000 (0.0000--0.0000) & 0.0000 (0.0000--0.0001) & 0.0000 (0.0000--0.0095) & 0.4258 (0.0118--0.5632) & 0.3107 (0.1504--0.4632) & --- \\
    Pancreas & 0.1491 (0.1079--0.2013) & 0.0000 (0.0000--0.0000) & 0.0000 (0.0000--0.0000) & 0.0000 (0.0000--0.0328) & 0.0514 (0.0274--0.1124) & \textbf{0.1974 (0.1651--0.2405)} & --- \\
    Spleen & 0.6136 (0.5192--0.6702) & 0.0000 (0.0000--0.0000) & 0.0000 (0.0000--0.0000) & 0.0000 (0.0000--0.1174) & 0.5020 (0.0279--0.7914) & \textbf{0.8422 (0.8062--0.8666)} & --- \\
    Stomach & 0.1558 (0.0799--0.2630) & 0.0000 (0.0000--0.0000) & 0.0000 (0.0000--0.0000) & 0.0000 (0.0000--0.0506) & 0.3144 (0.0384--0.6648) & \textbf{0.3256 (0.1678--0.5321)} & --- \\
    Bladder & 0.6414 (0.4268--0.7834) & 0.0000 (0.0000--0.0000) & 0.0000 (0.0000--0.0000) & 0.0000 (0.0000--0.1089) & \textbf{0.7514 (0.5654--0.8462)} & 0.6892 (0.5121--0.8084) & --- \\
    \midrule
    \multicolumn{8}{c}{\textbf{UMD-PETMR for Multi-Organ Segmentation}} \\ \midrule
    Aorta & 0.0779 (0.0514--0.1099) & 0.0000 (0.0000--0.0000) & 0.0000 (0.0000--0.0000) & \textbf{0.1146 (0.0445--0.1807)} & 0.0092 (0.0000--0.0834) & 0.0905 (0.0120--0.1754) & --- \\
    Brain & 0.7758 (0.7357--0.8196) & 0.0000 (0.0000--0.0000) & 0.0000 (0.0000--0.0000) & 0.2193 (0.0494--0.4368) & 0.9174 (0.8938--0.9309) & \textbf{0.9249 (0.9121--0.9340)} & --- \\
    Colon & 0.1159 (0.0773--0.1613) & 0.0000 (0.0000--0.0000) & 0.0000 (0.0000--0.0000) & 0.0236 (0.0075--0.0576) & 0.0309 (0.0092--0.0707) & \textbf{0.1900 (0.1206--0.2524)} & --- \\
    Esophagus & 0.0310 (0.0219--0.0420) & 0.0000 (0.0000--0.0000) & 0.0000 (0.0000--0.0000) & 0.0357 (0.0160--0.0607) & 0.0184 (0.0101--0.0447) & \textbf{0.1378 (0.0907--0.1749)} & --- \\
    Heart & 0.6832 (0.6051--0.7515) & 0.0000 (0.0000--0.0000) & 0.0000 (0.0000--0.0000) & 0.1728 (0.0505--0.4283) & 0.0348 (0.0043--0.1954) & 0.0348 (0.0043--0.1954) & --- \\
    Kidney-L & 0.5787 (0.5060--0.6476) & 0.0000 (0.0000--0.0000) & 0.0000 (0.0000--0.0000) & 0.0808 (0.0444--0.1433) & 0.0279 (0.0000--0.3344) & \textbf{0.7126 (0.6405--0.7678)} & --- \\
    Kidney-R & 0.5927 (0.5037--0.6671) & 0.0000 (0.0000--0.0000) & 0.0000 (0.0000--0.0000) & 0.0665 (0.0370--0.1056) & 0.0987 (0.0066--0.4267) & \textbf{0.7529 (0.6770--0.8014)} & --- \\
    Liver & 0.7004 (0.4837--0.7863) & 0.0000 (0.0000--0.0000) & 0.0721 (0.0573--0.0906) & 0.4072 (0.1121--0.5897) & 0.0117 (0.0032--0.0785) & \textbf{0.7966 (0.6880--0.8713)} & --- \\
    Lung & 0.1626 (0.0983--0.2570) & 0.0000 (0.0000--0.0000) & 0.0000 (0.0000--0.0001) & 0.0122 (0.0037--0.0585) & \textbf{0.2457 (0.0075--0.4651)} & 0.1883 (0.1375--0.2643) & --- \\
    Pancreas & 0.1727 (0.1112--0.2213) & 0.0000 (0.0000--0.0000) & 0.0000 (0.0000--0.0000) & 0.0385 (0.0154--0.0782) & 0.0255 (0.0038--0.0781) & \textbf{0.1963 (0.1257--0.2592)} & --- \\
    Spleen & 0.4270 (0.3100--0.5621) & 0.0000 (0.0000--0.0000) & 0.0000 (0.0000--0.0000) & 0.1215 (0.0554--0.2437) & 0.0258 (0.0018--0.1690) & \textbf{0.7119 (0.6021--0.8034)} & --- \\
    Stomach & 0.2747 (0.2007--0.3636) & 0.0000 (0.0000--0.0000) & 0.0000 (0.0000--0.0000) & 0.1012 (0.0460--0.1807) & 0.1203 (0.0751--0.2368) & \textbf{0.4749 (0.3970--0.5805)} & --- \\
    Bladder & 0.7219 (0.5580--0.8107) & 0.0000 (0.0000--0.0000) & 0.0000 (0.0000--0.0000) & 0.2201 (0.0783--0.4159) & 0.6536 (0.4636--0.7679) & \textbf{0.7944 (0.6538--0.8529)} & --- \\
    \midrule
    \multicolumn{8}{c}{\textbf{AutoPET-PSMA for Prostate Carcinoma Segmentation}} \\ \midrule
    PCa & 0.0924 (0.0166--0.2941) & 0.0525 (0.0035--0.2661) & 0.0000 (0.0000--0.0000) & 0.0230 (0.0037--0.0827) & 0.2222 (0.0470--0.4984) & 0.3427 (0.1120--0.5760) & \textbf{0.5759 (0.2005--0.8286)} \\
    \bottomrule
  \end{tabular}
  }
  \vspace{5pt}
  \caption{\textbf{Supplementary Table 4}: Quantitative comparison of DSC performance with other promptable segmentation foundation models and SegAnyPET on multi-organ, lymphoma, lung cancer, prostate carcinoma, UMD-CT organ and UMD-MRI organ tasks. The results are presented as the median Dice score along with the interquartile range (Q1--Q3). The best performing score for each target structure is highlighted in bold.}
  \label{tab:sp4}
\end{table}

    % =========================

\end{document}